\documentclass[10pt,twocolumn,letterpaper]{article}

\usepackage{wacv}
\usepackage{times}
\usepackage{epsfig}
\usepackage{graphicx}
\usepackage{amsmath}
\usepackage{amssymb}
\usepackage[noend]{algpseudocode}
\usepackage{algorithm}
\usepackage{subcaption}
\usepackage[noend]{algpseudocode}
\usepackage{algorithm}
\usepackage{lipsum}
\def\wacvPaperID{713} 

\wacvfinalcopy 
\ifwacvfinal
\def\assignedStartPage{9876} 
\fi

\newcommand\blfootnote[1]{%
  \begingroup
  \renewcommand\thefootnote{}\footnote{#1}%
  \addtocounter{footnote}{-1}%
  \endgroup
}

\ifwacvfinal
\usepackage[breaklinks=true,bookmarks=false]{hyperref}
\else
\usepackage[pagebackref=true,breaklinks=true,colorlinks,bookmarks=false]{hyperref}
\fi

\ifwacvfinal
\else
\fi

\begin{document}

\title{Towards Zero-Shot Learning with Fewer Seen Class Examples}

\author{
Vinay Kumar Verma{$^{1*}$}\hspace{1.2cm}Ashish Mishra{$^{2*}$}\hspace{1.2cm}Anubha Pandey{$^{2\$}$}\\
Hema A. Murthy{$^2$} \hspace{1.2cm}Piyush Rai{$^1$}\\
$^1$Department of CSE, IIT Kanpur; $^2$Department of CSE, IIT Madras,\\
{\tt\small \{vkverma,piyush\}@cse.iitk.ac.in; \{mishra,hema\}@cse.iitm.ac.in; anubhap93@gmail.com } 
}










\maketitle

\begin{abstract}
    We present a meta-learning based generative model for zero-shot learning (ZSL) towards a challenging setting when the number of training examples from each \emph{seen} class is very few. This setup contrasts with the conventional ZSL approaches, where training typically assumes the availability of a sufficiently large number of training examples from each of the seen classes.  The proposed approach leverages meta-learning to train a deep generative model that integrates variational autoencoder and generative adversarial networks. We propose a novel task distribution where meta-train and meta-validation classes are disjoint to simulate the ZSL behaviour in training. Once trained, the model can generate synthetic examples from seen and unseen classes. Synthesize samples can then be used to train the ZSL framework in a supervised manner. The meta-learner enables our model to generates high-fidelity samples using only a small number of training examples from seen classes.  We conduct extensive experiments and ablation studies on four benchmark datasets of ZSL and observe that the proposed model outperforms state-of-the-art approaches by a significant margin when the number of examples per seen class is very small.
\end{abstract}

\vspace{-2em}
\section{Introduction}
\blfootnote{$^\ast$Equal contribution. $\$$ Currently author is affiliated with MasterCard}
The traditional machine learning models for supervised classification assume the availability of labeled training examples from all the classes. This requirement is unrealistic and is rarely true in many real-world classification problems that consist of an ever-growing set of classes. Zero-Shot Learning (ZSL)~\cite{DEVISE,cmt} is a learning paradigm aimed at addressing this issue and learns to predict labels of inputs from the previously unseen classes. In particular, ZSL assumes that, during training, we have access to labeled examples from a set of \emph{seen} classes, and the test examples come from \emph{unseen} classes, i.e., classes that were not present during training. The ZSL methods typically accomplish learning using the class attributes/descriptions of the seen and unseen classes that help bridge the two. 
    
Recent work on ZSL has shown the effectiveness of deep learning-based approaches \cite{cmt,dem,lampert2014attribute,Chen_2018_CVPR,vermageneralized,f-CLSWGAN,liu2018generalized,changpinyo2016synthesized,CVPR19visual,CVPR19invarient,CVPR19latent,wang2017zero}. However, these methods usually assume that a large number of labeled example from each of the seen classes are available to train the model. In practice, it may often be challenging and time-consuming to collect a large number of labeled examples from the seen classes; sometimes, very few (say 5-10) labeled examples per seen classes may be available. Recently, meta-learning based frameworks \cite{maml,vinyals2016matching,ravi2016optimization} have addressed the issue of labeled data scarcity and have shown promising results for supervised few-shot learning problems\cite{ICCV19few2,ICCV10few1,CVPR19few3,CVPR19few2,CVPR19few1}. The basic idea in meta-learning is to train a model on various learning tasks, such that it can solve a new learning task using only a small number of training samples from each task. However, the meta-learning framework cannot handle the scenario if no labeled data is available for the unseen classes (new tasks). In this work, we propose incorporating the ZSL setup in meta-learning to handle this issue.
    
The recent works on ZSL, generative models have gained a significant attention  \cite{vermageneralized,f-CLSWGAN,Chen_2018_CVPR,mishra,verma2019meta} and have shown promising results for ZSL as well as for the more challenging setting of ZSL called \emph{generalized} zero-shot learning (GZSL).  In GZSL, the test inputs may be from seen as well as unseen classes. The success of generative models for ZSL leverages over recent advances in deep generative models, such as VAE \cite{VAE} and GAN \cite{GAN,wgan}. These generative models~\cite{vermageneralized,f-CLSWGAN,Chen_2018_CVPR,mishra,verma2019meta} can generate the \emph{synthetic} examples from unseen and seen classes (given their class attributes), using which we can train a classifier, which essentially turns the ZSL problem into a supervised learning problem. The generative approach is appealing but suffers from several shortcomings: $(i)$ These models do not mimic the ZSL behavior in training; therefore, there may be a large gap between the original and generated samples' quality. $(ii)$ These methods require a significant amount of labeled data, and $(iii)$ GAN based models generate high-quality samples but suffer from mode collapse. In contrast, VAE based models generate diverse samples, but there is a large gap between generated samples and the actual samples.
    
We develop a meta-learning-based ZSL framework that integrates a conditional VAE (CVAE) and a conditional GAN (CGAN) architecture to address the issues mentioned above. We divide the dataset into task/episode where, unlike standard meta-learning, each task has \emph{disjoint} support-set and query-set classes. The disjoint classes help to mimic the ZSL behaviour during training. The joint architecture helps to overcome the mode collapse problem and generates diverse samples for the seen and unseen classes. The augmentation of the joint architecture of VAE-GAN, with the meta-learning model, helps to train the model even when very few (say 5 or 10) samples per seen class are available. Once trained, the model can synthesize unseen class samples. These synthesized samples can be used to train any supervised classifier, essentially turning ZSL into a supervised learning problem.  Our generation based framework is also ideal for the \emph{generalized} ZSL problem~\cite{liu2018generalized,vermageneralized,CVPR19smoothing,CVPR19rethinking,verma2019meta,CVPR19active} where the test inputs can be from seen as well as unseen classes. 

Developing a meta-learning approach for ZSL is challenging. In the supervised meta-learning setup~\cite{maml,vinyals2016matching,ravi2016optimization}, we have meta-train, meta-validation, and meta-test. The meta-train and meta-validation share the same classes, while meta-test classes are disjoint. Also, the meta-test contains a few labeled samples for each unseen class. In the ZSL setup for the meta-test, we do not have any labeled samples but have access only to the class attribute vector.
Contrary to the standard meta-learning approach, the meta-train and meta-validation classes are disjoint in the proposed ZSL setting. The disjoint class simulates the ZSL setup during training. For the task distribution of the data, we follow the N-way, K-shot setup proposed by \cite{vinyals2016matching} and trained our model under Model-Agnostic Meta-Learning (MAML)~\cite{maml} framework that adapted to the ZSL setting. The main contributions of the proposed approach are summarized as follows:
    \begin{itemize}
        \item We propose a meta-learning based framework for ZSL and generalized ZSL problem. We train the joint VAE-GAN model with a meta-learner's help to generate diverse, high-quality samples that yield superior ZSL prediction accuracy.
        \item We propose a novel task distribution different from the traditional meta-learning model \cite{maml,vinyals2016matching} that mimics the ZSL setup in training and helps generate robust unseen class samples.
        \item Our framework can learn even when only a few labeled examples (say 5 or 10) are available per seen class. On the standard datasets, our approach shows state-of-the-art results against the strong generative baselines for the ZSL and GZSL.
    \end{itemize}

     \section{Related Work}
	\vspace{-7pt}
    Some of the earliest approaches for ZSL are based on learning a mapping from visual space to semantic space. During the test time, they use the learned mapping to predict the class-attributes for unseen examples and then do a nearest neighbor search to predict the class label \cite{lampert2014attribute,norouzi2013zero,cmt}. Similarly, some methods also use transfer learning between seen classes and unseen classes. They represent the parameters of unseen classes as a similarity weighted combination of the parameters of seen classes \cite{ESZSL,dem}. All these approaches rely on the availability of plenty of labeled data from the seen classes and do not perform well in the GZSL setting.
    
    Another popular approach to ZSL focuses on learning the linear/bilinear compatibility function between the visual and semantic spaces. Methods, such as ALE~\cite{ALE}, DEVISE~\cite{DEVISE}, ESZSL~\cite{ESZSL},\cite{relation}, learn a relationship between the inputs and class-attributes via a linear compatibility function, whereas SAE~\cite{SAE} adds an autoencoder loss to the projection that encourages re-constructability from the attribute space to the visual space. Some of the approaches~\cite{song2018transductive,xu2017transductive,CVPR19paul} assume that all test inputs of unseen classes are present during training and use these \emph{unlabeled} examples also in training. This setting is referred to as \emph{transductive} setting because of the extra information model leads to improve performance. The transductive assumption is often not realistic since the test inputs are usually not available during training. The attention-based approach is also explored in \cite{ji2018stacked,zhu2019semantic}, and these approaches work well for the fine-grain datasets. Paper \cite{hu2018correction,liu2018generalized} are calibration based approach and \cite{hu2018correction} use metric based meta-learning for the feature calibration.
	
    The Generalized ZSL (GZSL) problem is a more realistic and challenging problem as compared to standard ZSL. Unlike standard ZSL, in GZSL, the seen (training) and unseen (test) classes are not disjoint, which makes ZSL harder since the classifier is biased towards classifying each input (seen/unseen class) as belonging to the seen class. Most of the previous approaches perform well in standard ZSL but fail to handle biases towards seen classes~\cite{ALE,ESZSL,SAE,DEVISE}. Recently, generative models have shown promising results for both ZSL and GZSL setting. These approaches synthesize examples from both seen and unseen classes to train a supervised classifier, which somewhat mitigates the bias towards seen classes~\cite{GFZSL,Chen_2018_CVPR,mishra,cycle-consistancy,f-CLSWGAN,CVPR19feature,CVPR19dualgan,CVPR19paul,CVPR19alignedvae}. Most of the recent generative models for ZSL are based on VAE~\cite{CVAE} and GAN~\cite{GAN}. Among the generative models, ~\cite{mishra,vermageneralized,CVPR19alignedvae,wang2017zero, SBIR-CVPRW, SBIR-ECCV} are based on VAE architectures, while \cite{yu2019zero,ni2019dual,f-CLSWGAN,cycle-consistancy,Chen_2018_CVPR,CVPR19dualgan,verma2019meta, SBIR-WACV, SBIR-ICCVW, ZSAR-Neu, VinayWACV2020} use adversarial learning for sample generation based on the class-attribute. 
    The current generative approaches are unable to mimic ZSL behaviour in training. This limits the model's performance; also, they require significant labeled data from seen classes.
    In contrast, our proposed model Meta-VGAN leverage on the meta-learning framework can easily learn using very few labeled examples (as few as 5-10) from the seen classes. Also, disjoint task distribution helps to mimic ZSL behaviour in training. Our proposed model mainly focuses on ZSL/GZSL when very few labeled examples from seen classes are available. In contrast, we also conduct experiments using all examples from seen classes and observe that the proposed approach outperforms all recent state-of-the-art methods.
    
    	\begin{figure*}[thb]
		\centering
		\includegraphics[scale=0.42]{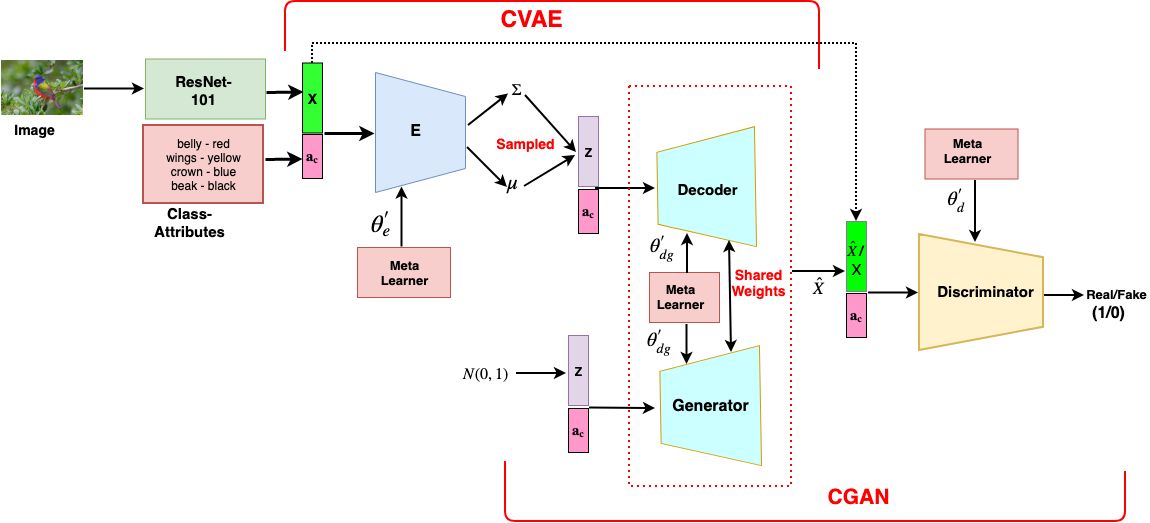}
		\caption{The overall architecture for our Meta-VGAN Zero Shot Learning model}
		\label{cvaegan}
		\vspace{-10pt}
	\end{figure*}
    
    \section{Proposed Model}
    \subsection{Problem Setup}
    \vspace{-6pt}
    \label{sec:pset}
    In ZSL, we assume that classes are divided into a disjoint set of seen (train) and unseen (test) classes. Only seen class examples are present during training, while the test set contains examples form unseen classes. Let $S$ and $U$ denote the number of seen/train and unseen/test classes, respectively. In ZSL, we also assume having access to class-attribute vectors/descriptions $\mathcal{A}=\{\mathbf{a_c}\}_{c=1}^{S+U}$, $\mathbf{a_c}\in \mathbb{R}^d$ for each of the seen and unseen classes. Our ZSL approach is based on synthesizing labeled features from each (unseen/seen) class using the trained generative model given the class attribute. Later, synthesized label features can be used to train a classifier; essentially, this turns the ZSL problem to a standard supervised learning problem.
    
    \textbf{Task-Distribution:} We use the MAML \cite{maml} as the base meta-learner. Note that unlike few-shot learning, in ZSL, no labeled data are available for the unseen classes. Therefore we also need to modify the episode-wise training scheme used in traditional MAML. We have train and test samples for ZSL where train and test classes are disjoint, also each class are associated with an attribute vector. The training data further divide into train-set and validation-set. Meta-learning further divides the train-set into a set of tasks following the N-way and K-shot setting, where each task contains N classes and K samples for each class. In particular, let $P(\mathcal{T})$ be the distribution of tasks over the train-set i.e. $P(\mathcal{T})=\{\tau_1,\tau_2,\dots, \tau_n\}$, $\tau_i$ is a task and $\tau_i=\{\tau_i^{v},\tau_i^{tr}\}$. The set $\tau_i^v$ and  $\tau_i^{tr}$ are in the $N$-way $K$-shot setting. However, unlike the traditional meta-learning setup, the classes of $\tau_i^v$ and  $\tau_i^{tr}$  are \emph{disjoint}. The disjoint class split simulate the ZSL setting during training and helps to generate high-quality unseen class samples.
	
\subsection{Motivation for Zero-Shot Meta-Learning}
    The standard meta-learning \cite{maml,vinyals2016matching,snell2017prototypical,ravi2016optimization} can quickly adapt to novel tasks using only a few samples. As demonstrated in prior work~\cite{vinyals2016matching,snell2017prototypical}, even without any labeled data form the novel classes, the meta-learning models can learn robust discriminative features. These works have also shown that sometimes zero-gradient step (without fine-tuning) model performs better than the fine-tuned architecture (refer to Table 1 in \cite{vinyals2016matching}). Another recent work \cite{raghu2019rapid_feature_reuse} demonstrates that the meta-learning framework is more about feature reuse instead of quick adaption. They have shown that a trained meta-learning model can provide a good enough feature for the novel classes that secure high classification accuracy without any fine-tuning. This motivates to a train of a generative model (VAE, GAN) in the meta-learning framework to synthesize the novel class samples. Our approach is similar in spirit to \cite{raghu2019rapid_feature_reuse}. The main difference is that ($i$) the proposed approach learns a generative model instead of discriminative, and ($ii$) Our problem setting is ZSL as opposed to a few-shot classification. In particular, the proposed approach's objective is to solve ZSL by generating novel class samples, while the goal in~\cite{raghu2019rapid_feature_reuse} is to learn discriminative features that are suitable for (few-shot) classification.
\subsection{Meta Learning-based Generative Model}
The proposed generative model (Meta-VGAN), shown in Fig.~\ref{cvaegan}, is a combination of conditional VAE (CVAE) and conditional GAN (CGAN). The model has shared parameters between the decoder and the GAN generator. The decoder's reconstructed samples are also fed to the discriminator; this increases the decoder's generation robustness. The decoder implicitly also works as a classifier. The Meta-VGAN consists of four modules - Encoder ($E$), decoder ($De$), Generator ($G$), and Discriminator ($Ds$). Each of these modules is augmented with a meta-learner. The $E$ and $De$ modules are from CVAE, whereas $G$ and $Ds$ modules are formed CGAN. Here $De$ and $G$ share common network parameters over different inputs. Unlike other generative models for ZSL \cite{f-CLSWGAN, cycle-consistancy}, we do not need a classifier over the $G$ module, since $De$ ensure the separability of the different classes generated by $G$. Let the parameters of the module $E$ and $Ds$ be denoted by $\theta_e$ and $\theta_{d}$ and the shared parameter of $De$ and $G$ be denoted by $\theta_g$.
In Fig.-\ref{cvaegan}, we have shown the module-wise overall structure of the proposed model. The objective function of CVAE is defined as:
    \begin{equation}\label{eq:vae1}
    \begin{split}
    \mathcal{L}^{V}(\theta_e,\theta_{g}) = & -KL(q_{\theta_e}(\mathbf{z|x,a_c})||p(\mathbf{z|a_c})) \\ & +\mathrm{E}_{\mathbf{z} \sim q_{\theta_e}(\mathbf{z|x,a_c})}[\log p_{\theta_{g}}(\mathbf{x|z,a_c})]
      \end{split}
    \end{equation}
    Here $\mathbf{a_c}\in \mathbb{R}^k$ is the attribute vector of the class for the input $\mathbf{x}$ $\in$ $\tau^{tr}_i$, and $\tau^{tr}_i\in\tau_i$. The conditional distribution $q_{\theta_e}(\mathbf{z|x,a_c})$ is parametrized by the encoder $E(\mathbf{x,a_c},\theta_e)$ output, the prior $p(\mathbf{z|a_c})$ is assumed to be $N(0,\mathbf{I})$, and $p_{\theta_{g}}(\mathbf{x|z,\mathbf{a_c}})$ is parametrized by the decoder $De(\mathbf{z,a_c},\theta_g)$ output. Similarly, for CGAN, the discriminator objective $\mathcal{L}^{D}$ and generator objective $\mathcal{L}^{G}$ are
    \begin{equation}\label{eq:dis}
    \small
    \begin{split}
    \mathcal{L}^{D}(\theta_{g},\theta_{d}) = &\mathbb{E}[Ds(\mathbf{x, a_c}|\theta_{d})]  -\mathbb{E}_{\mathbf{z}\sim N(0,1)}[Ds(G(\mathbf{z,a_c}|\theta_{g}),\mathbf{a_c}|\theta_{d})] \\& -\mathbb{E}_{\mathbf{z}\sim N(0,1)}[Ds(De(\mathbf{z,a_c},\theta_g)),\mathbf{a_c}|\theta_{d})]
    \end{split}
    \end{equation}
    \begin{equation}\label{eq:g}
    \mathcal{L}^{G}(\theta_{g},\theta_{d}) = \mathbb{E}_{\mathbf{z}\sim N(0,\mathbf{I})}[Ds(G(\mathbf{z,a_c}|\theta_{g}),\mathbf{a_c}|\theta_{d})]
    \end{equation}
    Next, we adapt the above objective to the MAML framework over the proposed task distribution (Sec.~\ref{sec:pset}). Our approach samples a task $\tau_i$ from the task distribution $P(\mathcal{T})$ i.e. $\tau_i\sim P(\mathcal{T})$ where $\tau_i=\{\tau_i^{v},\tau_i^{tr}\}$. The $\tau_i^{tr}$ is used to train the inner loop of the joint objective and $\tau_i^{v}$ is used to update the meta-objective. The joint objective of the CVAE and generator $G$ for the task $\tau_i$ is defined as:
    \vspace{-0.5em}
    \begin{equation}\label{eq:5}
    l^{VG}_{\tau_i}= \min_{\theta_{eg}} \left(-\mathcal{L}^{V}(\theta_e,\theta_{g})+    \mathcal{L}^{G}(\theta_{g},\theta_{d})\right)
    \end{equation}
    where $\theta_{eg}=[\theta_e,\theta_g]$ collectively denotes all the parameters of $E$ and $G$/$De$.
    The discriminator $Ds$ considers the samples generated from the $G$ and $De$ as fake samples and samples obtained from $\tau_i$ as real samples. Its objective for each task $\tau_i$ is given as:
    \begin{equation}\label{eq:6}
    l^D_{\tau_i}=\max_{\theta_d}\mathcal{L}^{D}(\theta_{g},\theta_{d})
    \end{equation}
    In Eq.~\ref{eq:5}-\ref{eq:6} $\tau_i$ is $\tau_i^{tr}$ for the inner loop and $\tau_i^v$ for the global update. A brief description is provided below.  
    The meta-learner (inner loop) update are performed on $\tau^{tr}\in \tau_i$. Empirically, for the ZSL setting, which is considerably harder than standard meta-learning, we observe that updates over individual tasks are unstable during training (because of GAN). To solve this problem, unlike the standard meta-learning model where updates are performed on each task, we update the meta-learner over a batch of tasks, i.e., the loss is averaged out over the set of tasks. Therefore, our inner loop objective is given by:
    \vspace{-0.5em}
    \begin{equation}\label{eq:7}
    \begin{split}
    & \min_{\theta_{eg}}\sum_{\tau_{i}^{tr} \in \tau_i \sim P(\mathcal{T})} l_{\tau_{i}^{tr}}^{VG} \hspace{2pt} \ \ \text{and} \ \ \hspace{2pt} \max_{\theta_{d}}\sum_{\tau_{i}^{tr} \in \tau_i \sim P(\mathcal{T})}     l^D_{\tau_{i}^{tr}}
    \end{split}
    \end{equation}
    The inner loop update for Eq.~\ref{eq:5}, i.e. joint objective of CVAE and $G$ is performed as:
    \begin{equation}\label{eq:eg}
    \theta_{eg}'\leftarrow     \theta_{eg}-\eta_1\nabla_{\theta_{eg}}\sum_{\tau_{i}^{tr} \in \tau_i \sim P(\mathcal{T})} l_{\tau_{i}^{tr}}^{VG}(\theta_{eg})
    \end{equation}
    Similarly the inner loop update for Eq.~\ref{eq:6} i.e. $Ds$'s objective is performed as:
    \vspace{-0.5em}
    \begin{equation}\label{eq:d}
    \theta_{d}'\leftarrow     \theta_{d}+\eta_2\nabla_{\theta_{d}}\sum_{\tau_{i}^{tr} \in \tau_i \sim P(\mathcal{T})} l_{\tau_{i}^{tr}}^{D}(\theta_{d})
    \end{equation}
    The optimal parameters obtained by meta-learner ($\theta_{eg}'$ and $\theta_d'$) are further applied on the meta-validation set $\tau_i^v$. Note that, since classes of $\tau_i^v$ and $\tau_i^{tr}$ are disjoint, in the outer loop the optimal parameters $\theta_{eg}'$ and $\theta_d'$ are applied to \emph{novel} classes. With these parameters, if the model can generate novel class samples that fool $Ds$, it indicates that the model has the ability to generate novel class samples. Otherwise, the outer loop is updated, and $\theta_{eg}'$ and $\theta_d'$ are considered the initializer for the model on the outer loop. The outer loop objective for the Eq.~\ref{eq:5} and Eq.~\ref{eq:6} is given by:
    \begin{equation}\label{eq:10}
    \small
    \begin{split}
    \min_{\theta_{eg}}\sum_{\tau_{i}^{v} \in \tau_i \sim P(\mathcal{T})} l_{\tau_{i}^{v}}^{VG}(\theta_{eg}') \quad \&  \quad \max_{\theta_{d}}\sum_{\tau_{i}^{v} \in \tau_i \sim P(\mathcal{T})} l_{\tau_{i}^{v}}^{D}(\theta_{d}')
    \end{split}
    \end{equation}
    The global update (outer loop) for the Eq.~\ref{eq:10} is:
    \vspace{-0.5em}
    \begin{equation}\label{eq:egf}
    \theta_{eg}\leftarrow     \theta_{eg}-\eta_1\nabla_{\theta_{eg}}\sum_{\tau_{i}^{v} \in \tau_i \sim P(\mathcal{T})} l_{\tau_{i}^{v}}^{VG}(\theta_{eg}')  
     \end{equation}
     \begin{equation}\label{eq:df}
    \theta_{d}\leftarrow     \theta_{d}+\eta_2\nabla_{\theta_{d}}\sum_{\tau_{i}^{v} \in \tau_i \sim P(\mathcal{T})} l_{\tau_{i}^{v}}^{D}(\theta_{d}')
    \end{equation}
    
    The algorithm \ref{alg:algorithm1} represents the training procedure of the proposed Meta-VGAN model.
\subsection{Zero-Shot Classification using Synthesized Examples}
\vspace{-0.5em}
Once the Meta-VGAN model is trained, we synthesize the examples (both unseen or seen classes) using their respective class attribute vectors passed to the generator/decoder module ($G_{\theta_dg})$. The generation of unseen class examples is done as $\mathbf{\hat{x}}= G_{\theta_dg}(\mathbf{z,a_c})$. Here $\mathbf{z}$ is sampled from a unit Gaussian i.e. $\mathbf{z} \sim N(0,\mathbf{I})$. The sampled $\mathbf{z}$ is concatenated with the class-attributes $\mathbf{a_c}$ and passed to the decoder as input, and it generates $\mathbf{\hat{x}}$, i.e., feature vectors of input from class $\mathbf{c}$. These synthesized unseen class examples can then be used as labeled examples to train any supervised classifier (e.g., SVM/Softmax). In the GZSL setting, we synthesize \emph{both} the seen and the unseen class examples using the class attributes and train an SVM or softmax classifier on all these training examples. The Experiments section contains further details of our overall procedure. 

 \begin{algorithm}[!ht]
  \caption{Meta-VGAN for ZSL}\label{alg:algorithm1}
  \begin{algorithmic}[1]
    \Require $p(\mathcal{T})$:distribution over tasks
    \Require $\eta_1,\eta_2$: step-size hyperparameters
    \State Randomly initialize $\theta_{e},\theta_{g}, \theta_{d}$
    \While{not done}
        \State Sample batch of tasks $\mathcal{T}_i \sim p(\mathcal{T})$;
             where, $\mathcal{T}_i=\{\mathcal{T}_{i}^{tr},\mathcal{T}_{i}^{v}\}$ such that $\mathcal{T}_{i}^{tr} \cap \mathcal{T}_{i}^{v} = \phi$
      \ForAll{$\mathcal{T}_i$}
        \State Evaluate $\nabla_{\theta_{eg}} l_{\mathcal{T}_{i}^{tr}}^{VG}(\theta_{eg})$
         \State Evaluate $\nabla_{\theta_{d}} l_{\mathcal{T}_{i}^{tr}}^{D}(\theta_{d})$
        \State Compute adapted parameters: $\theta_{eg}^{'} = \theta_{ed} - \eta_1\nabla_{\theta_{eg}} l_{\mathcal{T}_{i}^{tr}}^{VG}(\theta_{eg})$
        \State Compute adapted parameters: $\theta_{d}^{'} = \theta_{d} + \eta_2\nabla_{\theta_{d}} l_{\mathcal{T}_{i}^{tr}}^{D}(\theta_{d})$
      \EndFor
    \State Update $\theta_{eg} \leftarrow \theta_{eg} - \eta_1 \nabla_{\theta_{eg}} \sum_{\mathcal{T}_i \sim p(\mathcal{T})} l_{\mathcal{T}_{i}^{v}}^{VG}(\theta_{eg}^{'})$
    \State Update $\theta_{d} \leftarrow \theta_{d} + \eta_2 \nabla_{\theta_{d}} \sum_{\mathcal{T}_i \sim p(\mathcal{T})} l_{\mathcal{T}_{i}^{v}}^{D}(\theta_{d}^{'})$
    \EndWhile
  \end{algorithmic}
\end{algorithm}

\vspace{-1em}

\begin{table}[htb!]
\centering
\addtolength{\tabcolsep}{-1.5pt}
\begin{tabular}{l|c|c|c}
\hline 
Dataset & Attribute/Dim & \#Image & Seen/Unseen Class \\ \hline \hline
AwA2\cite{xian2018zero}    & A/85          & 37322   & 40/10             \\ \hline
CUB\cite{CUB}     & CR/1024       & 11788   & 150/50            \\ \hline
SUN\cite{SUN}     & A/102         & 14340   & 645/72            \\ \hline
aPY\cite{aPY}     & A/64          & 15339   & 20/12             \\ \hline
\end{tabular}
\vspace{-5pt}
\caption{The benchmark datasets used in our experiments, and their statistics.}
\label{tab:datasets}
\vspace{-10pt}
\end{table}
\vspace{-1em}
\section{Experiments}
	
    We conduct experiments on four widely used benchmark datasets and compare our approach with several state-of-the-art methods. Our datasets consist of Animals with Attributes (AwA)~\cite{xian2018zero}, aPascal and aYahoo (aPY)~\cite{aPY} , Caltech-UCSD Birds-200-2011 (CUB-
    200)~\cite{CUB} and SUN Attribute (SUN-A)~\cite{SUN}. The Table \ref{tab:datasets} shows description about the datsets.The details description of the datasets are provided in the supplementary material. We report our results on ZSL evaluation metrics proposed by~\cite{xian2018zero}. In particular, for GZSL, the harmonic mean of the seen and unseen class accuracies is reported, while for ZSL, the mean of the per-class accuracy is reported.
	These evaluation metrics helps to evaluate the model in an unbiased way. ResNet-101 features are used for all the datasets. All the baselines models also use the same features and evaluation metrics. 
	Due to space limitations, further details of the implementation and experimental setting are provided in the supplementary material.

	\subsection{Comparison with ZSL baselines}
	To prove our proposed model's efficacy, we create a baseline for a few examples per seen class using various recent state-of-the-art models. We follow the same experimental settings and the same number of samples for all the baseline as we used in the proposed model. 
	\begin{itemize}
		\item \textbf{CVAE-ZSL~\cite{mishra}}: CVAE-ZSL is a conditional variational autoencoder based generative model converts a ZSL problem into a typically supervised classification problem by using synthesized labeled inputs for the unseen classes. This approach generates the samples conditioned on the unseen class attributes, using a CVAE. The generated samples are used to train a classifier for the unseen classes. For the GZSL setting, both seen and unseen class samples are generated, and the classifier is trained on the union set.
		\item \textbf{GF-ZSL}~\cite{vermageneralized} GF-ZSL, a state-of-the-art method, is an extended version of the CVAE-ZSL model, where it uses a feedback mechanism in which a discriminator (a multivariate regressor) learns to map the generated samples to the corresponding class attribute vectors, leading to an improved generator.
		\item \textbf{f-CLSWGAN~\cite{f-CLSWGAN}} f-CLSWGAN uses the Wasserstein-GAN as a generative model with an additional classifier associated with the generator. The classifier increases the class separation between two classes, while the generator's goal is to generate samples that follow the original data distribution closely.
		\item \textbf{cycle-UWGAN~\cite{cycle-consistancy}} This approach uses cycle consistency loss as a regularizer with Wasserstein-GAN, which helps to reconstruct back semantic features from generated visual features. Using cycle consistency loss, it synthesizes more representative visual features for both seen and unseen classes. 
		\item \textbf{ZSML~\cite{verma2019meta}} Recently ZSML~\cite{verma2019meta} train the GAN in the meta-learning framework, that helps to generate the robust samples for the novel classes. The proposed approach shows a significant improvement over the ZSML model.
	\end{itemize}

	\begin{table}[h]
	\vspace{-5pt}
		\begin{center}
			\scalebox{0.8}{
				\addtolength{\tabcolsep}{3.0pt}
				\begin{tabular}{l|l|c|c|c|c}
					\hline 
					\textbf{Method} &  \textbf{N} & \textbf{SUN} & \textbf{CUB} & \textbf{AwA2} &  \textbf{APY} \tabularnewline
					\hline 
					\hline
					{CVAE-ZSL \cite{mishra}} &5 & 49.7 & 52.8 & 41.5 & 21.6 \\& 10 & 51.2 & 53.5 & 45.6 & 21.8\\ 
					\hline
					{GF-ZSL \cite{vermageneralized}} &5 & 53.2 & 56.5 & 57.4 & 32.6 \\& 10 & 55.3 &56.9 & 59.2 & 35.1\\ 
					\hline
					{cycle-UWGAN \cite{cycle-consistancy}} &5 & 45.6 & 56.8 & 61.9 & 36.3\\& 10 & 45.8 & 57.2 & 62.0 & 38.1\\ 
					\hline
					{f-CLSWGAN \cite{f-CLSWGAN}} &5 & 32.6 & 40.9 & 59.8 & 41.4 \\& 10 & 34.5 & 41.5 & 62.3 & 41.9\\ 
					\hline \hline
					{\textbf{Ours(Meta-VGAN)}}&5 &\textbf{59.1} & \textbf{66.9} & \textbf{67.3} & \textbf{48.6} \\& 10&\textbf{60.3} & \textbf{68.8} & \textbf{68.4} & \textbf{49.2}\\ 
					\hline 
				\end{tabular} 
			}
			\vspace{-2pt}
			\caption{Zero-Shot learning (ZSL) results only using five and ten example per seen classes to train the model}
			\label{tab:zsl_f}
			\vspace{-2em}
		\end{center}
	\end{table}		

\vspace{-0.5em}
	\subsection{Results}
	 We train our model and all the baselines using five and ten examples per seen class. For this experiment, we compare our model with CVAE-ZSL, GF-ZSL, f-CLSWGAN, and cycle-UWGAN. We consider both the ZSL (test examples only from unseen classes) as well as generalized ZSL (test examples from seen as well as unseen classes) settings. To prove our proposed Meta-VGAN model's efficacy, we also perform the experiments using all the examples from seen classes and compare against recent methods (refer Table-\ref{tab:a}). 
	\vspace{-.5em}
	\subsection*{Zero-Shot Learning (ZSL)}
	For the ZSL setting, we compare our Meta-VGAN with the state-of-the-art methods on all the datasets in Table \ref{tab:zsl_f}. The proposed approach shows a significant improvement over the baseline methods for all four datasets. We use a small training set (5 and 10 samples per seen class) to train the model in all of these experiments. We observe that our Meta-VGAN model can train the model well using only a few examples per seen class and significantly better results than the baselines. The proposed approach achieves the substantial absolute performance gains by more than $11.1\%$, $5.4\%$, $7.2\%$, and $6.4\%$, in comparison with the baseline methods, on CUB, AWA2, aPY, and SUN datasets, respectively for five examples per seen class.
	\subsection*{Generalized Zero-Shot Learning (GZSL)}
	    In the GZSL settings, the test input is classified into the joint class space of the seen and unseen classes, i.e., $Y = Y^{S} \cup Y^{ U}$. The GZSL setting is more practical as it removes the assumption that test input only comes from unseen classes. For the evaluation metric, we compute the harmonic mean \cite{xian2018zero} of seen and unseen class accuracy's: $H=2*S * U/(S +U)$, where $S$ and $U$ denote the accuracy of seen classes and unseen classes respectively. We evaluate our method on three standard datasets and show the performance comparison with the baseline approach in Table \ref{tab:gzsl_f}.
    
    To have a fair comparison with the previous state-of-the-art generative methods, we keep the number of synthetic samples per class the same. For the baselines,  we follow the same setup as in the original papers and perform experiments in our novel setup when only a few samples per seen class are available. We perform the experiments in two settings. In the first setting, we assume that each seen class has only five samples for training, while in the other setting, we assume that ten samples are available. In both cases, our Meta-VGAN method shows the improvements by notable margins on all the datasets compared to all baseline methods. Our model achieves the substantial absolute performance gains in the harmonic mean by $10.8\%$, $11.9\%$, and $3.4\%$ in comparison with the baseline method on CUB, AWA2, and aPY datasets, respectively.
	\vspace{-5pt}

	\section{Ablation Study}
	\vspace{-0.5em}
    To disentangle the role of each component of the proposed framework, we conduct a detailed ablation analysis. We are primarily interested in the following: ($i$) The performance with and without meta-learner. ($ii$) The performance in traditional ZSL when the number of examples per seen class is \emph{not} very small. ($iii$) Investigating why meta-learner helps in improved sample generation and which of the components of our model benefit more from meta-learning and ($iv$) The effect of disjoint task distribution vs standard task distribution.

\begin{table*}[h]
		\begin{center}
			\scalebox{0.9}{
				\addtolength{\tabcolsep}{7.2pt}
				\begin{tabular}{|l|l|l|l|l|l|l|l|l|l|l|}
					\hline
					{\textbf{Method}} & {\textbf{N}}& \multicolumn{3}{c|}{\textbf{AwA2}} & \multicolumn{3}{c|}{\textbf{CUB}} &
					\multicolumn{3}{c|}{\textbf{aPY}} \\ \cline{3-11} 
					& & \textbf{U} & \textbf{S} & \textbf{H} & \textbf{U} & \textbf{S} & \textbf{H} &  \textbf{U} & \textbf{S} & \textbf{H} \\ \hline \hline 
					{CVAE-ZSL \cite{mishra}}&5 &16.5 & 28.7 & 21.0 & 50.1 & 28.1 & 36.0 &  19.9 & 65.4 & 30.5\\ &10 & 18.1 & 29.2 & 22.3 & 50.5 & 28.7 & 36.5 &  20.5 & 65.8 & 31.2 \\ \hline
					{cycle-UWGAN \cite{cycle-consistancy}}&5 & 40.4 & 43.3 & 41.8 & 48.2 & 33.3 & 39.4 &  18.6 & 64.2 & 28.8 \\ &10 & 45.5 & 50.9 & 48.0 & 48.3 & 35.2 & 40.7 &  19.6 & 66.3 & 30.2 \\ 
					\hline
					{f-CLSWGAN \cite{f-CLSWGAN}} & 5& 37.8 & 44.2 & 40.7 & 30.4 & 28.5 & 29.4 &  17.2 & 40.6 & 24.5 \\ &10 & 40.5 & 55.9 & 46.9 & 34.7 & 38.9 & 36.6 &  21.2 & 32.2 & 25.6\\ 
					\hline
					{GF-ZSL \cite{vermageneralized}} &5 & 38.2 & 44.3 & 41.0 & 29.4 & 33.0 & 31.0 &  20.2 & 66.3 & 30.9\\ &10 & 41.4 & 45.9 & 43.5 & 35.6 & 43.5 & 39.1 &  20.6 & 67.8 & 31.5 \\
					\hline
					{ZSML \cite{verma2019meta}} &5 & 38.4 &61.3 &47.3 & 32.9 &38.2 &35.3 &  -- & -- & --\\ &10 & 47.8 &59.6 &53.1 & 42.7 &45.1 &43.9 &  -- & -- & -- \\
					\hline
					\hline
					{\textbf{Ours (Meta-VGAN)}}& 5 & 44.5  & 67.5  &\textbf{53.7}  &{51.5}  & 47.8  &\textbf{50.2}  &{24.1}  &58.9  & \textbf{34.3} \\ &10 & 46.2 & 73.1 & \textbf{56.6} & 52.5 & 49.2 & \textbf{52.1} &  24.5 & 59.1 & \textbf{35.3} \\
					\hline
				\end{tabular}
			}
			\vspace{-5pt}
			\caption{GZSL results when only five and ten examples per seen class are used to train the model. We randomly selected 5/10 samples per class, for examples in the CUB dataset for 150 training classes we have new dataset size 150*(5/10), rest samples are not used.}
			\label{tab:gzsl_f}
			\vspace{-2em}
		\end{center}
	\end{table*}

 \subsection{With and without meta-learning }
    \vspace{-0.5em}
    The proposed model trains the generative model in the meta-learning framework. To illustrate the contribution of the meta-learner module, we perform an experiment when no meta-learning component is present. We trained the model with and without meta-learner, for CUB and AWA2 datasets, for standard ZSL and GZSL settings. We observe that a meta-learner's role in our proposed model is crucial, which helps to train the model very efficiently using a few examples per seen class. The meta-learner component boosts the model's absolute performance by $7.8\%$ and $6.4\%$ in standard ZSL, while by $9.9\%$ $6.2\%$ in GZSL for CUB and AWA2 datasets, respectively. Please refer to supplementary material for more detail.
    \subsection{What if we DO have plenty of training examples from seen classes?}
    \vspace{-0.5em}
    Our meta-learning-based model is primarily designed for the setting when the number of training examples per seen class is very small (e.g., 5-10). We also investigate whether meta-learning helps when the number of examples per seen class is \emph{not} small (i.e., the conventional ZSL setting). In this experiment, we consider the complete dataset for our experiment; this is essentially a traditional ZSL setting.
    
    We conduct experiments on both standard ZSL and GZSL setting for CUB and AWA2 datasets, and the results are shown in Table \ref{tab:a}. Using all the examples from seen classes to train the model, we observe that our model improves the result with a significant margin compared to all baseline approaches in both the settings. For the ZSL setting, our model achieves $9.4\%$ and $2.1\%$ improvement as compared to the state-of-the-art result on CUB and AWA2 datasets, respectively. Similarly, for the GZSL setting, the model achieves consistent performance gain harmonic mean (a more meaningful metric) on CUB and AWA2 datasets. We observe that all existing baseline methods show a significant difference in performance between the two regimes, i.e., using all samples and using only a few samples. In contrast, our proposed model shows competitive results in both cases. It shows that the existing baseline approaches are not suitable when the number of seen class examples is very small.

\begin{table*}[h]
		\begin{center}
			\scalebox{0.9}{
				\addtolength{\tabcolsep}{11.2pt}
				\begin{tabular}{lcccccccc}
					\hline
					\multicolumn{1}{|c|}{{\textbf{Method}}} & \multicolumn{2}{c|}{\textbf{Standard Setting}} & \multicolumn{6}{c|}{\textbf{Generalized Setting}} \\ \cline{2-9} 
					\multicolumn{1}{|c|}{} & \multicolumn{1}{c|}{{\textbf{CUB}}} & \multicolumn{1}{c|}{{\textbf{AwA2}}} & \multicolumn{3}{c|}{\textbf{CUB}} & \multicolumn{3}{c|}{\textbf{AwA2}} \\ \cline{4-9} 
					\multicolumn{1}{|c|}{} & \multicolumn{1}{c|}{} & \multicolumn{1}{c|}{} & \multicolumn{1}{c|}{\textbf{U}} & \multicolumn{1}{c|}{\textbf{S}} & \multicolumn{1}{c|}{\textbf{H}} & \multicolumn{1}{c|}{\textbf{U}} & \multicolumn{1}{c|}{\textbf{S}} & \multicolumn{1}{c|}{\textbf{H}} \\ \hline 
					\multicolumn{1}{|l|}{CVAE-ZSL \cite{mishra}} & \multicolumn{1}{c|}{52.1} & \multicolumn{1}{c|}{65.8} & \multicolumn{1}{c|}{-} & \multicolumn{1}{c|}{-} & \multicolumn{1}{c|}{34.5} & \multicolumn{1}{c|}{-} & \multicolumn{1}{c|}{-} & \multicolumn{1}{c|}{51.2} \\ \hline
					\multicolumn{1}{|l|}{cycle-UWGAN\cite{cycle-consistancy}} & \multicolumn{1}{c|}{58.6} & \multicolumn{1}{c|}{66.8} & \multicolumn{1}{c|}{47.9} & \multicolumn{1}{c|}{59.3} & \multicolumn{1}{c|}{53.0} & \multicolumn{1}{c|}{59.6} & \multicolumn{1}{c|}{63.4} & \multicolumn{1}{c|}{59.8} \\ \hline
					\multicolumn{1}{|l|}{f-CLSWGAN \cite{f-CLSWGAN}} & \multicolumn{1}{c|}{57.3} & \multicolumn{1}{c|}{68.2} & \multicolumn{1}{c|}{43.7} & \multicolumn{1}{c|}{57.7} & \multicolumn{1}{c|}{49.7} & \multicolumn{1}{c|}{57.9} & \multicolumn{1}{c|}{61.4} & \multicolumn{1}{c|}{59.6} \\ \hline
					\multicolumn{1}{|l|}{GF-ZSL \cite{vermageneralized}} & \multicolumn{1}{c|}{59.6} & \multicolumn{1}{c|}{69.2} & \multicolumn{1}{c|}{41.5} & \multicolumn{1}{c|}{53.3} & \multicolumn{1}{c|}{46.7} & \multicolumn{1}{c|}{{58.3}} & \multicolumn{1}{c|}{68.1} & \multicolumn{1}{c|}{{62.8}} \\ \hline \hline
					\multicolumn{1}{|l|}{\textbf{Ours (Meta-VGAN)}} & \multicolumn{1}{c|}{\textbf{70.4}} & \multicolumn{1}{c|}{\textbf{73.2}} & \multicolumn{1}{c|}{{55.2}} & \multicolumn{1}{c|}{48.0} & \multicolumn{1}{c|}{\textbf{53.2}} & \multicolumn{1}{c|}{57.4} & \multicolumn{1}{c|}{70.5} & \multicolumn{1}{c|}{\textbf{63.5}} \\ \hline
				\end{tabular}
			}
			\vspace{-0.5pt}
			\caption{The comparison of our ZSL and GZSL result with the recent state-of-the-art generative model when using all samples. All the approach follow the same setting proposed by \cite{xian2018zero}}
			\label{tab:a}
			\vspace{-2em}
		\end{center}
	\end{table*}

\begin{figure*}[h!]
\begin{subfigure}{.5\textwidth}
  \centering
  \includegraphics[width=.7\linewidth]{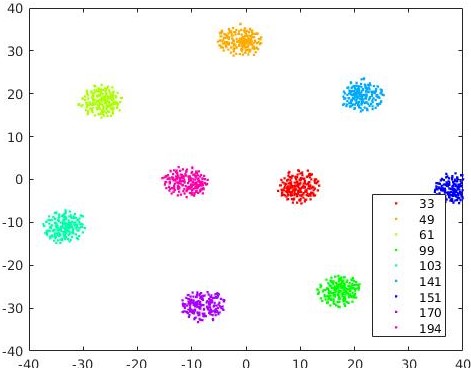}  
  \caption{Use all samples from seen classes to train the model}
  \label{fig:sub-first}
\end{subfigure}
\begin{subfigure}{.5\textwidth}
  \centering
  \includegraphics[width=.7\linewidth]{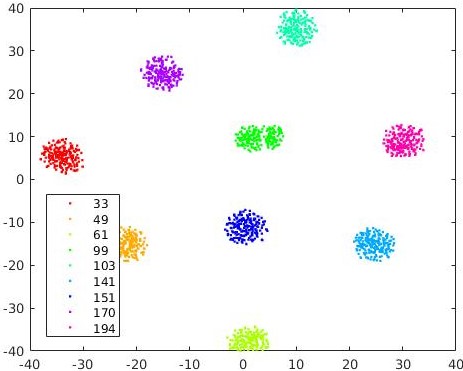}  
  \caption{Using 5 examples per seen class to train the model}
  \label{fig:sub-second}
\end{subfigure}
\vspace{-1em}
\caption{t-NSE visualization of synthesized features on the CUB dataset for the random 10 classes. We can observe that using very few training samples the generated data distribution are very close to the case when model use all samples. }
\label{fig:fig}
\end{figure*}

	\subsection{Why meta-learning helps and which model components benefit by it the most?}

    The meta-learning framework is capable of learning using only a few samples from each seen class. The episode-wise training helps the meta-learning framework to adapt to the new task quickly using only a few samples. In the zero-shot setup, we do not have the samples of the unseen classes; therefore, quick adaption is not possible in this case. In our proposed approach, we trained the model such that adaption is based solely on the class attribute vectors. Later, the unseen class attribute vectors can be used for unseen class sample generation.  In the case of adversarial training, the generation quality depends on the quality of discriminator, generator, and the feedback provided by discriminator to the generator. The inner loop of the meta-learner provides reliable guidance (initialization) to the outer loop. As we know, adversarial training is sensitive to initialization, in our case, because of the better initialization model learns the robust parameter. The guided discriminator network is capable of learning the optimal parameters with the help of a very few examples. Therefore discriminator provides strong feedback to the generator. The guided (better initialized) generator uses this strong feedback, and because of its improved power, the generator can counter the discriminator. The alternating optimization with the guided generator and discriminator improves the overall learning capability of the adversarial network. Similarly, the meta-learning framework also helps the VAE architecture using the inner loop guidance to initialize the parameter. Therefore the joint framework in the proposed approach improves the generation quality even though only very few samples are available per seen class. 
    
    We also experiment with the discriminator without meta-learning. On CUB, this reduces the ZSL accuracy from $70.4\%$ to $65.6\%$. The performance drop occurs because, without meta-learning, the discriminator is sub-optimal and is unable to provide robust feedback to the generator. Therefore the generator is unable to learn robustly. Also, we observe that removal of meta-learner from the generator (but not the discriminator), over CUB dataset, the performance drops from $70.4\%$ to $68.8\%$. Through this analysis, we observe that the meta-learning based discriminator is more critical than the meta-learning based generator.
\subsection{Disjoint task distribution vs. standard task distribution}
The empirical evaluation shows that the proposed disjoint task distribution helps to generate robust samples for the seen/unseen classes. We have disjoint classes between $\tau_i^{tr}$ and $\tau_i^v$. Therefore, if the inner loop parameter is optimal only for the $\tau_i^{tr}$, it produces a high loss for the $\tau_i^v$. Consequently, the network tries to learn an optimal parameter that minimizes the loss on the meta-train as well as the meta-validation set. The evaluation of the proposed split vs. standard split supports our claim. On the CUB and AWA2 dataset for the ZSL setup, the proposed split shows the $68.8\%$ and  $68.4\%$, while the standard split shows poor performance, and we obtained $66.2\%$ and $65.7\%$ mean per-class accuracy.
\subsection{A simpler generative model trained with meta-learner  for ZSL}
We also compare with vanilla CVAE trained with meta-learning (a simple generative model) with the Meta-VGAN. The results are shown in Table~\ref{tab:cvaefewshot}, we observe that a complex model (CVAE+CGAN) synthesizes better quality features for unseen classes as compared to a simple generative model. The proposed model outperforms over CVAE based model for all datasets by a significant margin. Therefore the joint model has better generalization ability for the feature generation of the unseen classes.

\begin{table}[h]
    \centering
    \addtolength{\tabcolsep}{18pt}
    \begin{tabular}{l l l }
    \hline
    {\textbf{Datasets}} & \multicolumn{2}{c}{\textbf{Accuracy}} \\ \cline{2-3} & \textbf{5}   & \textbf{10}   \\  \hline \hline
    SUN \cite{SUN} & 54.95 & 56.54  \\ \hline
    CUB\cite{CUB} & 63.70 & 65.63 \\ \hline
    AWA1 \cite{xian2018zero} & 64.20 & 64.32 \\ \hline
    AwA2\cite{xian2018zero} & 64.22 & 64.85  \\ \hline
    aPY & 42.15 & 42.99 \\ \hline
    \end{tabular}
    \caption{ZSL results using vanilla CVAE as generative model trained with meta learner over the five standard dataset.}
\label{tab:cvaefewshot}
\vspace{-2em}
\end{table}

\section{Conclusion}
We proposed a novel framework to solve ZSL/GZSL when very few samples from each of the seen classes are available during training. The various components in the proposed joint architecture of the conditional VAE and conditional GAN are augmented with meta-learners. The GAN helps to generate high-quality samples while VAE reduces the mode collapse problem that is very common in GANs. The meta-learning framework with episodic training requires only a few samples to train the complete model. The meta-learners inner loop provides a better initialization to the generative structure. Therefore, the guided discriminator provides strong feedback to the generator, and the generator can generate high-quality samples. In the ablation study, we have shown that meta-learning based training and disjoint meta-train and meta-validation classes are the crucial components of the proposed model. Extensive experiments over benchmark datasets for ZSL/GZSL show that the proposed model is significantly better than the baselines.

{\small
\bibliographystyle{ieee_fullname}
\bibliography{egbib}
}

\newpage
\newpage
\appendix
\section{Algorithm}
 \begin{algorithm}[!ht]
  \caption{Meta-VGAN for ZSL}\label{alg:algorithm1}
  \begin{algorithmic}[1]
    \Require $p(\mathcal{T})$:distribution over tasks
    \Require $\eta_1,\eta_2$: step-size hyperparameters
    \State Randomly initialize $\theta_{e},\theta_{g}, \theta_{d}$
    \While{not done}
        \State Sample batch of tasks $\mathcal{T}_i \sim p(\mathcal{T})$;
             where, $\mathcal{T}_i=\{\mathcal{T}_{i}^{tr},\mathcal{T}_{i}^{v}\}$ such that $\mathcal{T}_{i}^{tr} \cap \mathcal{T}_{i}^{v} = \phi$
      \ForAll{$\mathcal{T}_i$}
        \State Evaluate $\nabla_{\theta_{eg}} l_{\mathcal{T}_{i}^{tr}}^{VG}(\theta_{eg})$
         \State Evaluate $\nabla_{\theta_{d}} l_{\mathcal{T}_{i}^{tr}}^{D}(\theta_{d})$
        \State Compute adapted parameters: $\theta_{eg}^{'} = \theta_{ed} - \eta_1\nabla_{\theta_{eg}} l_{\mathcal{T}_{i}^{tr}}^{VG}(\theta_{eg})$
        \State Compute adapted parameters: $\theta_{d}^{'} = \theta_{d} + \eta_2\nabla_{\theta_{d}} l_{\mathcal{T}_{i}^{tr}}^{D}(\theta_{d})$
      \EndFor
    \State Update $\theta_{eg} \leftarrow \theta_{eg} - \eta_1 \nabla_{\theta_{eg}} \sum_{\mathcal{T}_i \sim p(\mathcal{T})} l_{\mathcal{T}_{i}^{v}}^{VG}(\theta_{eg}^{'})$
    \State Update $\theta_{d} \leftarrow \theta_{d} + \eta_2 \nabla_{\theta_{d}} \sum_{\mathcal{T}_i \sim p(\mathcal{T})} l_{\mathcal{T}_{i}^{v}}^{D}(\theta_{d}^{'})$
    \EndWhile
  \end{algorithmic}
\end{algorithm}

    \begin{figure*}[h!]
		\begin{centering}
			\includegraphics[scale=0.36]{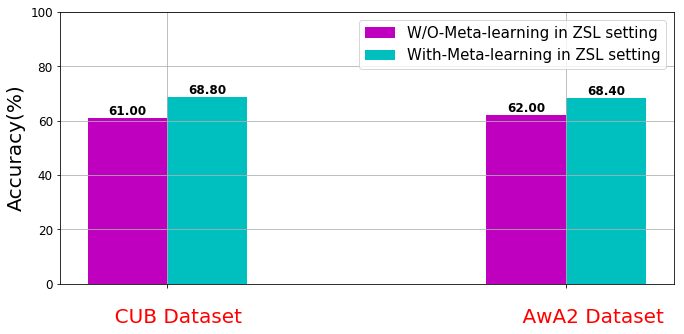}
			\includegraphics[scale=0.36]{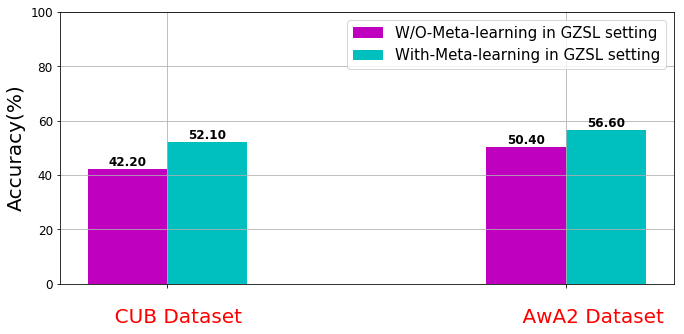}
			\caption{The left figure shows the mean per class accuracy with and without meta-learner in the ZSL setting. In the right figure the GZSL result are shown with and without meta-learning.} 
			\label{ab1}
		\end{centering}
	\end{figure*}
	
\section{Datasets Descriptions}

\begin{table}[htb!]
\centering
\addtolength{\tabcolsep}{-1pt}
\begin{tabular}{|l|c|c|c|}
\hline 
Dataset & Attribute/Dim & \#Image & Seen/Unseen Class \\ \hline \hline
AwA2\cite{AWA2}    & A/85          & 37322   & 40/10             \\ \hline
CUB\cite{CUB}     & CR/1024       & 11788   & 150/50            \\ \hline
SUN\cite{SUN}     & A/102         & 14340   & 645/72            \\ \hline
aPY\cite{aPY}     & A/64          & 15339   & 20/12             \\ \hline
\end{tabular}
\vspace{5pt}
\caption{The benchmark datasets used in our experiments, and their statistics.}
\label{tab:datasets}
\vspace{-10pt}
\end{table}

To evaluate our proposed model in comparison with several state-of-the-art ZSL and generalized ZSL methods, we applied our approach to the following benchmark ZSL datasets: SUN\cite{SUN}, CUB\cite{CUB}, AwA2\cite{AWA2}, and aPY \cite{aPY}. Table \ref{tab:datasets} shows the summary of the datasets used and their statistics.\\
\textbf{SUN Scene Recognition:} SUN is a fine-grained dataset with 717 scene categories and 14,340 images. We use the widely used split of the dataset for the ZSL setting, 645 seen classes, and 72 unseen classes. The dataset has image-level attributes. For training, we use class-level attributes obtained by combining the attributes of all the images in a class.\\
\textbf{Animals with Attributes:} AwA2 is a coarse-grained dataset with 50 classes and 37,322 images. We follow a standard zero-shot split of 40 seen (train) classes and ten unseen (test) classes. The dataset has 85-dimensional human-annotated class-attributes. \\
\textbf{Caltech UCSD Birds 200:} CUB is a fine-grained dataset with 11,788 images from 200 different types of birds, annotated with 312 attributes. We use a zero-shot split of 150 unseen and 50 seen classes. The dataset has image-level attributes like the SUN dataset. We average these image-level attributes of all the classes to obtain class attributes for training.\\
\textbf{Attribute Pascal and Yahoo (aPY):} aPY is a coarse-grained dataset with 64 attributes. The dataset has 32 classes. For Zero-Shot learning, we follow a split of 20 Pascal classes for training and 12 Yahoo classes for testing.

\section{Implementation Details}
Our proposed architecture Meta-VGAN has an encoder, decoder, generator, and discriminator modules, as shown in Figure-1 in the main paper. The decoder and generator modules share the same network parameters. Each of the modules has a series of FC layers followed by a ReLU and dropout layers. We concatenate image feature vector $\mathbf{x}$ extracted from ResNet-101 with class attributes vector $\mathbf{a}$ and feed to the Encoder module $E$. The encoder $E$ has a series of three FC layers, and encodes the input to $\mathbf{d_z}$ (varies with datasets) dimensional latent space with mean $\mathbf{\mu}$ and variance $\mathbf{\Sigma}$. Noise dimensions used for CUB, SUN, AwA2, and aPY datasets are 512, 20, 40, and 20, respectively. Next, we sample $\mathbf{d_z}$ dimensional noise vector from the latent space and feed it to the decoder module (or the generator module). The decoder (or generator) uses a series of 2 FC layers followed by ReLU to generate a 2048 dimension feature vector $\mathbf{\hat{x}}$ similar to the input image feature vector $\mathbf{x}$. The generated image features $\mathbf{\hat{x}}$ are further passed through the discriminator module. The discriminator receives two types of inputs: the real image feature vector $\mathbf{x}$ that comes from the ground truth data of the training set and the synthesized image features $\mathbf{\hat{x}}$ generated by the generator module(or the decoder module). The discriminator has a series of 3 FC layers and tries to distinguish between the real image feature vector $\mathbf{x}$ and the generated image feature vector $\mathbf{\hat{x}}$. The discriminator outputs the probability of the image is real. The output value should be close to 0 for fake image features $\mathbf{\hat{x}}$, and it should be close to 1 for real image features $\mathbf{x}$.

 	For training, we associate each of the modules with a meta-learner agent in the Meta-VGAN model. We randomly sample 10 classes for training and ten classes for validation from the seen classes of the dataset such that they are mutually exclusive. We call this a \emph{task}. We randomly sample 5 examples from each class of the train set and three examples from each class of the validation set. For each task, we iterate through each class of the train set multiple times and compute the adapted parameters of the network using Eq.7 and 8 in the main paper. Next, we pass the validation data through the network with initial parameters and with the computed parameters and compute the loss. We finally update the network on the validation loss, as shown in Eq.10 and 11 in the main paper.
    The learning rate and dropout rate used for all the datasets are 0.001 and 0.3, respectively.  All the hyperparameters are selected using cross-validation.
    
    The values of hyper-parameters $\eta_1$ and $\eta_2$, used for computation of updated parameters on training loss, are empirically chosen  using a grid search in the range $[{1e-1},{1e-8}]$.

    \subsection{Comparison with and without meta-learning }
    Our model is a combination of CVAE and CGAN, which are integrated with a meta-learner. To illustrate the contribution of the meta-learner module, we perform an experiment when no meta-learning component is present. Figure \ref{ab1} shows the comparison between with and without meta-learner in our model, for CUB and AWA2 datasets, in both standard ZSL and GZSL settings. We observe that the role of a meta-learner in our proposed model is very crucial, which helps to train our model very efficiently using a few examples per seen class. The meta-learner component boosts the model's absolute performance by $7.8\%$ and $6.4\%$ in standard ZSL, while by $9.9\%$ $6.2\%$ in GZSL for CUB and AWA2 datasets, respectively.

\end{document}